\documentclass{article}

\usepackage[preprint]{neurips_2025}

\usepackage[utf8]{inputenc} 
\usepackage[T1]{fontenc}    
\usepackage{hyperref}       
\usepackage{url}            
\usepackage{booktabs}       
\usepackage{amsfonts}       
\usepackage{nicefrac}       
\usepackage{microtype}      
\usepackage{xcolor}         

\usepackage{graphicx}
\usepackage{amsmath}
\usepackage{enumitem}
\usepackage{amsmath} 
\usepackage{graphicx}
\usepackage{wrapfig}
\usepackage{booktabs}
\usepackage{caption}
\usepackage{array}
\usepackage{wrapfig}
\usepackage{subcaption}
\usepackage{pifont}
\usepackage{bm}

\usepackage{float} 
\usepackage{booktabs} 
\usepackage{multirow}
\title{P-4DGS: Predictive 4D Gaussian Splatting with 90$\times$ Compression}

\author{
  Henan Wang\quad Hanxin Zhu\quad Xinliang Gong\quad Tianyu He\quad Xin Li\footnotemark[1]\quad Zhibo Chen\footnotemark[1]\\
  University of Science and Technology of China\\
  \texttt{\{henanwang, hanxinzhu, qq0707\}@mail.ustc.edu.cn}\\
  \texttt{\{xin.li, chenzhibo\}@ustc.edu.cn}
}

\begin{document}

\renewcommand{\thefootnote}{\fnsymbol{footnote}}
\footnotetext[1]{Corresponding Authors}

\maketitle

\begin{abstract}
3D Gaussian Splatting (3DGS) has garnered significant attention due to its superior scene representation fidelity and real-time rendering performance, especially for dynamic 3D scene reconstruction (\textit{i.e.}, 4D reconstruction). However, despite achieving promising results, most existing algorithms overlook the substantial temporal and spatial redundancies inherent in dynamic scenes, leading to prohibitive memory consumption. To address this, we propose P-4DGS, a novel dynamic 3DGS representation for compact 4D scene modeling. Inspired by intra- and inter-frame prediction techniques commonly used in video compression, we first design a 3D anchor point-based spatial-temporal prediction module to fully exploit the spatial-temporal correlations across different 3D Gaussian primitives. Subsequently, we employ an adaptive quantization strategy combined with context-based entropy coding to further reduce the size of the 3D anchor points, thereby achieving enhanced compression efficiency. To evaluate the rate-distortion performance of our proposed P-4DGS in comparison with other dynamic 3DGS representations, we conduct extensive experiments on both synthetic and real-world datasets. Experimental results demonstrate that our approach achieves state-of-the-art reconstruction quality and the fastest rendering speed, with a remarkably low storage footprint (around \textbf{1MB} on average), achieving up to \textbf{40$\times$} and \textbf{90$\times$} compression on synthetic and real-world scenes, respectively.
\end{abstract}

\section{Introduction}

Recently, 4D Gaussian Splatting (\textit{i.e.}, dynamic 3DGS)~\cite{yang2024deformable,yang2023real} has emerged as a powerful paradigm for reconstructing dynamic 3D scenes, enabling high-fidelity modeling of appearance and motion while supporting real-time rendering and continuous free-viewpoint exploration.

To achieve this goal, existing methods~\cite{yang2024deformable,wu20244d} mainly adopt a static canonical space combined with a time-varying deformation field to model scene dynamics. In this formulation, the canonical space encodes the structural information of the scene and serves as a reference for deformation field. The deformation field captures the temporal evolution of each 3D Gaussian, tracking both spatial trajectories and attribute changes across frames. For example, D3DGS~\cite{yang2024deformable} employs a deformable MLP to learn the deformation of each canonical Gaussian. However, despite their success, existing 4D reconstruction approaches typically suffer from substantial storage costs, severely limiting their scalability and real-world deployment.

To better understand and mitigate such inefficiencies, we draw an analogy between dynamic 3D Gaussian representations and established techniques in video compression~\cite{sullivan2012overview}. Intuitively, deformation field-based 4DGS frameworks parallel the predictive structures used in video coding, where the canonical space serves as a reference frame and the deformation field functions analogously to motion vectors that describe inter-frame changes. This observation motivates us to ask: \textit{can we design spatial and temporal prediction structures for dynamic Gaussians, analogous to intra- and inter-prediction in video compression?}

Specifically, video codecs exploit spatial redundancy via intra-frame prediction, temporal redundancy via inter-frame prediction, and contextual redundancy through entropy models such as CABAC~\cite{marpe2003context}. Inspired by this, we propose an efficient dynamic 3DGS representation (\textit{i.e.}, P-4DGS) that jointly leverages temporal and spatial prediction. For spatial prediction, we adopt a 3D anchor point-based predictive structure to exploit the spatial correlations among different 3D Gaussians within the canonical space, where nearby Gaussians are predicted by a single anchor point to reduce the number of primitives. For temporal prediction, we utilize the deformation MLP to predict the deformation vector of each 3D Gaussian from the canonical space to specific time. Furthermore, we introduce adaptive quantization and a context-aware entropy model to enhance the compression efficiency of the canonical space's 3D Gaussians. 

We verify our method in multiple benchmarks including both real scenes and synthetic scenes~\cite{yan2023nerf,pumarola2021d}. Results show that our method can achieve better rendering quality while greatly reducing the size compared to existing 4D compression methods.

The main contributions of this paper can be summarized as:
\begin{itemize}[left=0pt]
\item We propose P-4DGS, a novel dynamic 3D Gaussian representation designed for compact 4D scene reconstruction.

\item Inspired by intra- and inter-prediction paradigm of video coding, we design a spatial-temporal prediction module to exploit redundancies within dynamic 3D Gaussians. We further introduce adaptive quantization and context-based entropy coding to facilitate more efficient coding of these dynamic 3D Gaussians.

\item Experimental results demonstrate that our method achieves high compression efficiency—requiring \textbf{less than 1 MB of storage with a 90× compression ratio}—while delivering improved rendering quality.

\end{itemize}

\section{Related Work}

\subsection{Dynamic 3D Representation}
3D Gaussian Splatting~\cite{kerbl20233d} provides an innovative representation model for novel view synthesis. With modest storage needs and after short training sessions, 3DGS has the ability to carry out real-time, high-fidelity view synthesis in large-scale scenes.
Several works~\cite{wang2025gflow,wu20244d,yang2024deformable,huang2024sc} have adapted 3DGS to facilitate the reconstruction of dynamic scenes.
On one hand, recent works~\cite{yang2023real,duan20244d,li2024spacetime,kratimenos2024dynmf,katsumata2024compact} directly train a collection of 4D Gaussians to represent static, dynamic, and transient elements in a scene. Nevertheless, these approaches necessitate a substantial quantity of Gaussians to create high-quality scene representations, leading to significant storage costs.
On the other hand, a range of alternative works aim to construct geometry and depict dynamics by collaboratively optimizing the Gaussian and deformation fields in the canonical space. For example, Deformable-3DGS~\cite{yang2024deformable} utilizes an MLP deformation network to represent the motion in dynamic scenes. Similarly, 4DHexPlane~\cite{wu20244d} leverages Hexplane~\cite{cao2023hexplane} to encode sparse data and then outputs the results via a multi-head Gaussian decoder, successfully enabling real-time rendering at high resolution.

Although both categories achieve decent rendering quality for dynamic scenes, they still demand substantial storage to handle the multi-dimensional attributes of millions of 3D Gaussians in canonical Gaussians. Based on this, we propose to perform 4D scene reconstruction on a more compact Gaussian representation, and simultaneously incorporate joint optimization with entropy encoding to achieve the compression of the 4D scene.

\subsection{Compact 3D Representation}
To reduce the substantial memory requirements of 3DGS, researchers have developed two main strategies.
The first category focuses on traditional compression techniques without altering the original 3DGS representation. These traditional compression methods cover vector quantization~\cite{fan2024lightgaussian,lee2024compact,navaneet2023compact3d,niedermayr2024compressed,papantonakis2024reducing,xie2024mesongs}, pruning redundant Gaussians~\cite{fan2024lightgaussian,lee2024compact,wang2024end,papantonakis2024reducing,zhang2024gaussianspa,liu2024maskgaussian}, implicit encoding of high-dimensional attributes~\cite{girish2024eagles,lee2024compact,wu2024implicit}, utilization of standardized compression pipelines~\cite{fan2024lightgaussian,morgenstern2024compact,wu2024implicit,lee2025compression}, and implementation of entropy constraint~\cite{wang2024end,zhan2025cat}.
The second category explores more compact Gaussian representations to mitigate storage challenges~\cite{lu2024scaffold,hamdi2024ges}. A prominent example is Scaffold-GS~\cite{lu2024scaffold}, which introduces a unique method by assigning learnable features to a sparse set of anchor points that predict attributes for a broader set of neighboring 3D Gaussians. Building on Scaffold-GS, subsequent research has focused on optimizing entropy coding methods by designing various types of entropy models to more efficiently estimate the distributions of Gaussian attributes~\cite{chen2024hac,liu2024compgs,wang2024contextgs,chen2024fast,liu2024hemgs,chen2025hac++}. 

However, adapting methods from both categories to the reconstruction of dynamic scenes may not be straightforward, as most 4D extensions require substantial architectural modifications to extend 3DGS for dynamic scene modeling. In our 4D compression work, we conduct dynamic scene reconstruction on a more compact 3DGS and incorporate an entropy model for joint optimization. This approach enables us to achieve a high compression ratio while maintaining high rendering quality and speed.

\subsection{Video Coding}
Video compression aims to reduce storage and transmission costs by exploiting spatial and temporal redundancies in video data. Standard codecs such as H.264/AVC~\cite{wiegand2003overview}, H.265/HEVC~\cite{sullivan2012overview}, and H.266/VVC~\cite{bross2021overview} achieve this through intra- and inter-prediction. Intra-prediction operates within a single frame by predicting a block of pixels based on the values of its neighboring reconstructed blocks. It uses various directional modes (e.g., vertical, horizontal, angular) to estimate the content, then encodes only the residual (difference) between the prediction and the actual block. Inter-prediction predicts the current frame using one or more reference frames. It involves motion estimation to find matching blocks in reference frames and motion compensation to generate a predicted block using motion vectors. The encoder then stores the residual and motion data instead of the raw block. These prediction modules are followed by transform coding, quantization, and entropy coding steps to further compress the residuals and motion information.
    
\begin{figure}[t]
  \centering
  \includegraphics[width=1.0\textwidth]{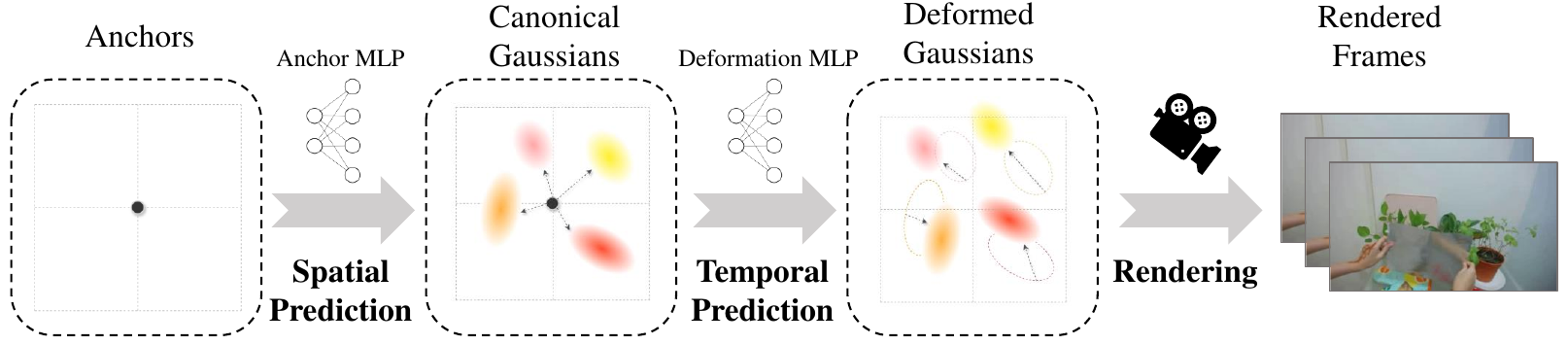}
  \caption{Rendering pipeline of P-4DGS. The pipeline first performs spatial prediction by mapping anchor points in the canonical space to static Gaussian primitives via an anchor prediction module. Then, temporal prediction is conducted using a deformation MLP that maps these primitives to a target time step $t$, producing dynamic Gaussian primitives for final image rendering.}
  \label{fig:render-pipeline}
  \vspace{-4mm}
\end{figure}

\section{Method}

\subsection{Overview}

\begin{figure}[t]
  \centering
  \includegraphics[width=1.0\textwidth]{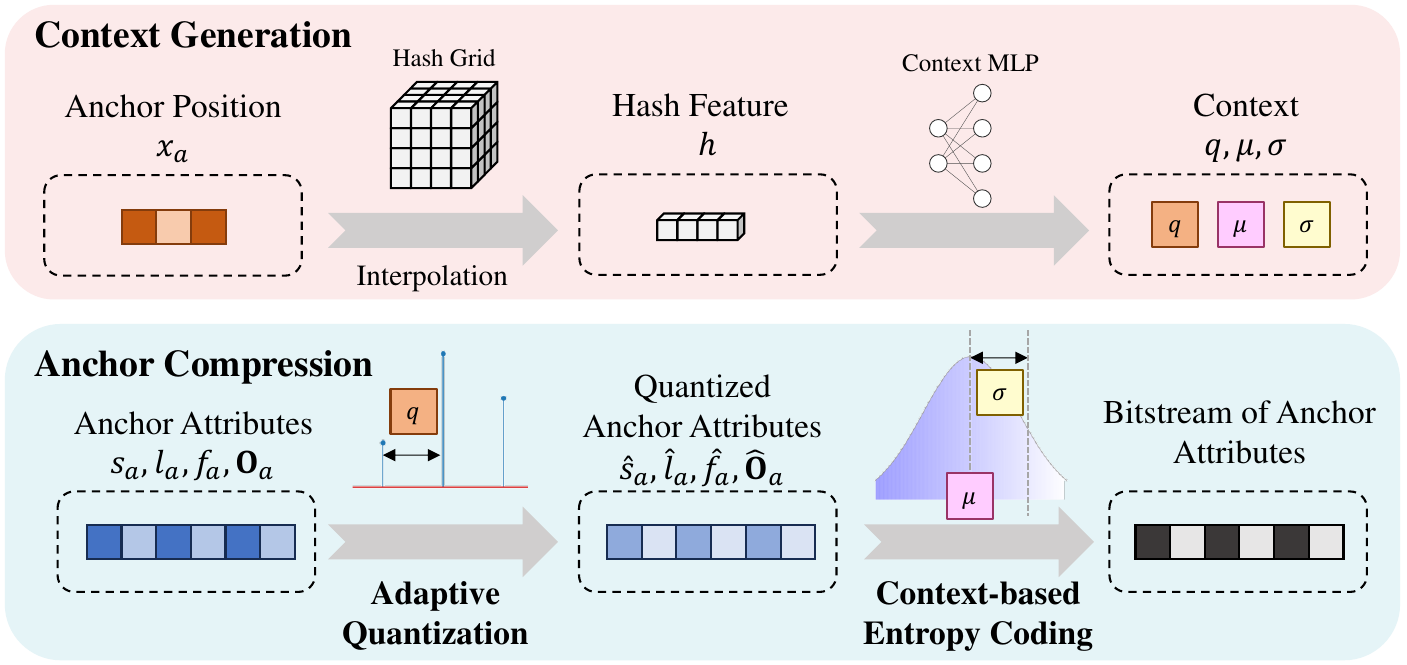}
  \caption{Entropy coding pipeline of P-4DGS, consisting of context generation and anchor compression. In context generation, anchor positions $x_a$ query a hash grid to produce a feature $h$. which an MLP maps to quantization step $q$, mean $\mu$, and standard deviation $\sigma$. In anchor compression, attributes $s,l,f,O$ are adaptively quantized using $q$ and encoded into a bitstream via context-based entropy coding with $\mu,\sigma$.}
  \label{fig:entropy_model}
\end{figure}

Fig.~\ref{fig:render-pipeline} and Fig.~\ref{fig:entropy_model} illustrate the rendering and entropy coding pipeline of our proposed P-4DGS, respectively. The rendering pipeline is responsible for the generation and rendering of dynamic 3D Gaussian primitives, while the entropy coding pipeline focuses on data compression.

In the rendering pipeline, anchor points in the canonical space are first processed through an anchor prediction module to generate static Gaussian primitives in the canonical space. These Gaussian primitives then serve as inputs to the temporal prediction module, where a deformation MLP is employed to compute the deformation field from the canonical space to the target time step $t$. The resulting deformation vectors are applied to the canonical Gaussian primitives to produce dynamic Gaussian primitives corresponding to time $t$, which are subsequently rendered into the final images through a rendering module.

In the entropy coding pipeline, the anchor points in the canonical space undergo quantization, followed by entropy coding to generate a compact bitstream representation. This process enhances the efficiency of both storage and transmission.

\subsection{Spatial-temporal Prediction}

\paragraph{Spatial prediction.}
The spatial prediction of Gaussian primitives primarily adopts an anchor-based prediction scheme~\cite{lu2024scaffold}, which exploits spatial correlations among Gaussian primitives. Several spatially adjacent Gaussian primitives are predicted from a single anchor, significantly reducing storage overhead. Each anchor is characterized by five key attributes: position $x_a \in \mathbb{R}^3$, scale $s_a \in \mathbb{R}^3$, offset scaling $l_a \in \mathbb{R}^3$, learnable offsets $\mathbf{O}_a \in \mathbb{R}^{k \times 3}$ and anchor feature $f_a \in \mathbb{R}^{d}$. 

During rendering, given a camera pose, each visible anchor within the view frustum generates $k$ nearby Gaussian primitives and predicts their corresponding attributes. Specifically, for an anchor located at $x_a$, the positions of its $k$ associated Gaussian primitives are computed as: 
\begin{equation}
    \left\{x_0,...,x_{k-1}\right\} = x_a + \left\{O_0,...O_{k-1}\right\} \cdot l_a,
\end{equation}
where $\left\{O_0,...O_{k-1}\right\}\in\mathbb{R}^{k\times3}$ are learnable offsets.

The remaining four attributes for the $k$ Gaussian primitives are predicted from the anchor feature $f_a$ by MLPs. These MLPs take as input the anchor feature $f_a$, the relative distance $\delta_{vc}$, and the viewing direction $d_{vc}$ between the anchor and the camera position:
\begin{equation}
    \delta_{vc} = \| x_a - x_c \|_2, \quad d_{vc} = \frac{x_a-x_c}{\| x_a - x_c \|_2}, 
\end{equation}
where $x_c$ denotes the camera position.

Using these inputs, the opacity of the $k$ Gaussian primitives is predicted as:
\begin{equation}
    \left\{\alpha_0,...,\alpha_{k-1}\right\} = \psi_{\alpha}\left(f_a, \delta_{vc}, d_{vc}\right),
\end{equation}
where $\psi_{\alpha}$ denotes the MLP for opacity.

The color $c$ and rotation $r$ attributes are predicted analogously. For the scale attribute, the MLP output is interpreted as a residual scaling factor relative to the anchor scale $s_a$:
\begin{equation}
    \left\{s_0,...,s_{k-1}\right\} = s_a \cdot \texttt{sigmoid}\left(\psi_s\left(f_a, \delta_{vc}, d_{vc}\right)\right),
\end{equation}
where $\psi_s$ denotes the MLP for scale attribute.

Like Scaffold-GS~\cite{lu2024scaffold}, to reduce rendering overhead, Gaussian primitives with opacity values below a threshold ($\alpha < \tau_\alpha$) are excluded from rendering.

\paragraph{Temporal prediction.}
Following D3DGS~\cite{yang2024deformable}, we adopt a canonical space plus deformation field model for temporal prediction. The set of Gaussian primitives predicted by anchors serves as the canonical space. Given a time input $t$, we query the deformation field for the corresponding deformation vectors at time $t$ and apply them to the Gaussian primitives in the canonical space, thereby obtaining the Gaussian primitives at time $t$.

Temporal prediction consists of two main components: positional encoding and the deformation MLP. Given the position of a Gaussian primitive and the time $t$, temporal prediction first encodes the spatial and temporal information into high-dimensional vectors, enhancing the MLP's ability to learn high-frequency variations. These encoded vectors are concatenated and fed into a deformation MLP to predict deformation vectors for the Gaussian primitive’s position, scale, and rotation, denoted as $(\Delta x, \Delta s, \Delta r)$:
\begin{equation}
    (\Delta x, \Delta s, \Delta r) = \psi_d(\texttt{concat}(\mathcal{E}(x), \mathcal{E}(t))),
\end{equation}
where \texttt{concat} indicates vector concatenation, $\mathcal{E}$ denotes positional encoding and $\psi_d$ is the deformation MLP. The deformed attributes of the Gaussian primitive at time $t$ are obtained by adding the deformation vectors to the canonical attributes:
\begin{equation}
    (x', s', r') = (x+\Delta x, s+\Delta s, r+\Delta r).
\end{equation}

\subsection{Adaptive Quantization and Entropy Coding}
\paragraph{Adaptive quantization.}
Anchor attributes, including scale $s_a \in \mathbb{R}^3$, offset scaling $l_a \in \mathbb{R}^3$, offsets of $k$ Gaussian primitives $\mathbf{O}_a \in \mathbb{R}^{k \times 3}$, and anchor features $f_a \in \mathbb{R}^{d}$, are quantized using scalar quantization after training to reduce storage overhead. To address the undifferentiable problem caused by quantization, we adopt uniform noise to simulate the quantization error during training~\cite{balle2016end}. Furthermore, the quantization step is adjusted adaptively for each attribute. Taking the anchor feature as an example, the adaptive quantization process during training is defined as:
\begin{equation}
    \tilde{f_a}=f_a + \mathcal{U}\left(-\frac{1}{2},\frac{1}{2}\right) \cdot q,
\end{equation}
where $\tilde{f_a}$ is the quantized anchor feature with added uniform noise, $\mathcal{U}(-\frac{1}{2}, \frac{1}{2})$ is a uniform distribution over $(-\frac{1}{2}, \frac{1}{2})$, and $q$ is the quantization step size, computed as:
\begin{equation}
    q = Q_0\cdot\left(1+\tanh(\psi_q(h))\right),
\end{equation}
where $Q_0$ is the base quantization step size, and $\psi_q$ is an MLP that learns a residual offset from $Q_0$ based on the context hash feature $h$. The quantization step size $q$ lies in the range $(0, 2Q_0)$. The hash feature $h$ is queried from a binary hash grid $\mathcal{H}$ using the anchor position $x_a$:
\begin{equation}
    h = \mathcal{H}(x_a),
\end{equation}
The hash grid consists of one 3D grid (along the $x$, $y$, $z$ dimensions) and three 2D grids (along the $xy$, $yz$, and $xz$ planes), where each grid point corresponds to an anchor in the canonical space and stores its contextual information.

During inference, hard quantization is performed using rounding:
\begin{equation}
    \hat{f_a} = \lfloor\frac{f_a}{q}\rceil \cdot q,
\end{equation}
where $\hat{f_a}$ is the quantized anchor feature.

\paragraph{Context entropy coding.}
To accurately estimate the bitrate and enable rate-distortion optimization, we design an entropy model to estimate the probability distribution of the quantized anchor attributes. We assume a Gaussian prior for the quantized values, following \cite{chen2024hac, balle2018variational}. Taking the anchor feature as an example, the discrete probability distribution is given by:
\begin{equation}
    \begin{split}
        p(\tilde{f}_a)&=\int_{\tilde{f}_a-\frac{1}{2}q}^{\tilde{f}_a+\frac{1}{2}q}\phi_{\mu,\sigma}(x)dx \\
        &= \Phi_{\mu,\sigma}\left(\tilde{f_a}+\frac{1}{2}q\right) - \Phi_{\mu,\sigma}\left(\tilde{f_a}-\frac{1}{2}q\right),
    \end{split}
\end{equation}
where $\phi_{\mu,\sigma}$, $\Phi_{\mu,\sigma}$ are the probability density function and cumulative distribution function of a Gaussian distribution with mean $\mu$ and standard deviation $\sigma$. These parameters are predicted by an MLP $\psi_p$ using the hash feature $h$:
\begin{equation}
    (\mu, \sigma) = \psi_p(h).
\end{equation}

Using this estimated distribution, the total bitrate for anchor encoding is computed as:
\begin{equation}
    R_{\rm anchor} = \sum_a (-\log_2 p(\tilde{f}_a) - \log_2 p(\tilde{\mathbf{O}}_a) - \log_2 p(\tilde{l}_a) - \log_2 p(\tilde{s}_a)).
\end{equation}

In addition, we also compute the bitrate required for encoding the binary hash grid. Let $p_+$ be the probability of a grid value being 1 and $N_+$ be the number of such values. Similarly, $p_- = 1 - p_+$ is the probability of a grid value being -1, and $N_-$ is its count. The hash grid bitrate is computed as:
\begin{equation}
    R_{\rm hash} = N_+(-\log_2(p_+)) + N_-(-\log_2(p_-)).
\end{equation}

\paragraph{Training objective.}
The final objective function combines the rendering loss (Eq.~\ref{eq: render_loss}) and the bitrate loss (including both anchor and hash grid bitrates), balanced by a rate weighting factor $\lambda_{\rm rate}$:
\begin{equation}
    \mathcal{L}_{\rm total}
    = \lambda_{\rm rate} \mathcal{L}_{\rm rate}
    + \mathcal{L}_{\rm render},
\end{equation}
\begin{equation}
    \label{eq: render_loss}
    \mathcal{L}_{\rm render} = \left(1-\lambda_{\rm SSIM}\right)\mathcal{L}_1 + \lambda_{\rm SSIM}\mathcal{L}_{\rm D-SSIM},
\end{equation}
\begin{equation}
    \mathcal{L}_{\rm rate} = R_{\rm anchor} + R_{\rm hash}.
\end{equation}
\begin{figure}[t]
    \centering
    \includegraphics[width=1.0\linewidth]{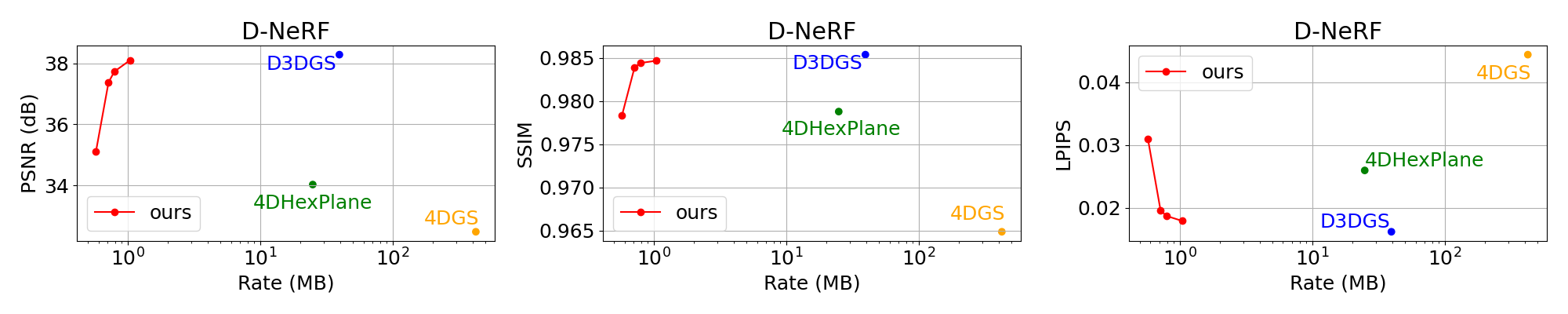}
    \includegraphics[width=1.0\linewidth]{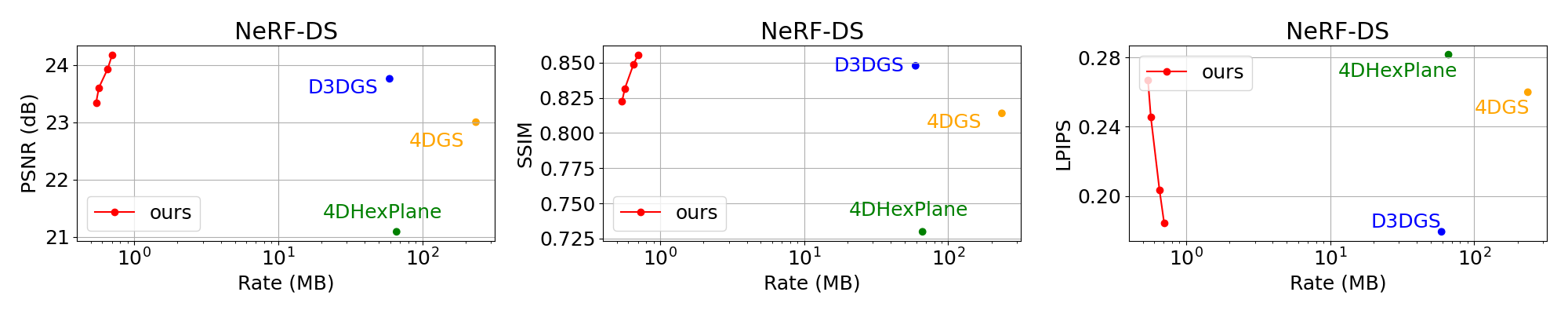}
    \caption{Rate-distortion curves on D-NeRF and NeRF-DS datasets. The x-axis shows the bitrate (log scale) of compressed Gaussian representations, and the y-axes report average PSNR, SSIM, and LPIPS. Our method achieves high reconstruction quality across bitrates, outperforming D3DGS, 4DHexPlane, and 4DGS, with over 40$\times$ and 90$\times$ compression on D-NeRF and NeRF-DS, respectively.}
    \label{fig:scaffold-4dgs_rd}
    \vspace{-4mm}
\end{figure}

\vspace{-4mm}
\section{Experiments}

\subsection{Experimental Settings}

\paragraph{Datasets.}
We evaluate the proposed method on two widely used dynamic 3D scene datasets: the synthetic D-NeRF dataset \cite{pumarola2021d} and the real-world NeRF-DS dataset \cite{yan2023nerf}. In the D-NeRF dataset, viewpoints are sampled on a trajectory centered around the object. In contrast, the NeRF-DS dataset consists of videos captured using a stereo camera setup with two fixed viewpoints. For both datasets, we follow the official train/test splits provided by the dataset authors.

\paragraph{Implementation details.}
The training of dynamic 3DGS, which combines temporal and spatial prediction, is divided into four main stages. In the first stage, a static canonical space is trained using all training images from different time steps, for a total of 3,000 iterations. In the second stage, quantization-aware training is introduced by injecting uniform noise into anchor point attributes to simulate quantization noise. Training continues up to 4,000 iterations for real scenes or 5,000 iterations for synthetic scenes. In the third stage, temporal information is incorporated, enabling the deformation MLP to learn mappings from the canonical space to specific time steps $t$ until 10,000 iterations. In the fourth stage, the entropy model is integrated into the training pipeline to estimate the bitrate of anchor attributes and enable joint rate-distortion optimization, continuing until 20,000 iterations for real scenes or 40,000 iterations for synthetic scenes.

\paragraph{Baselines.}
We compare our method against three representative dynamic 3D Gaussian representation approaches, including D3DGS~\cite{yang2024deformable}, 4DHexPlane~\cite{wu20244d}, and 4DGS~\cite{yang2023real}. Both D3DGS and 4DHexPlane adopt the canonical space–deformation field framework for temporal modeling. D3DGS employs the same type of deformation field as ours, consisting solely of a deformation MLP. 4DHexPlane, in addition to the deformation MLP, introduces a multi-resolution hex-plane representation to model spatiotemporal information. In contrast, 4DGS encodes temporal variation by augmenting each Gaussian with an explicit time attribute, resulting in a time-aware 4D Gaussian point cloud.

\begin{table}[t]
\small
  \caption{Rate-distortion performance and rendering FPS on D-NeRF dataset.}
  \label{tab:dnerf_results}
  \centering
   \begin{tabular}{@{}lcccccc@{}}
    \toprule
    Method & PSNR$\uparrow$ & SSIM$\uparrow$ & LPIPS$\downarrow$ & Rate (MB)$\downarrow$ & FPS$\uparrow$ \\
    \midrule
    D3DGS       & \textbf{38.28} & \textbf{0.985} & \textbf{0.016} & 39.45  & 149 \\
    4DHexPlane    & 34.02	& 0.984	& 0.021	& 23.45  & 132 \\
    4DGS        & 32.47	& 0.976	& 0.027	& 375.34 & 147 \\
    \midrule
    Ours        & 38.10	& \textbf{0.985}	& 0.017	& \textbf{1.039}  & \textbf{262} \\
  \bottomrule
  \end{tabular}
  \vspace{-4mm}
\end{table}

\begin{table}[t]
\small
  \caption{Rate-distortion performance and rendering FPS on NeRF-DS dataset.}
  \label{tab:nerfds_results}
  \centering
  \begin{tabular}{@{}lcccccc@{}}
    \toprule
    Method & PSNR↑ & SSIM↑ & LPIPS↓ & Rate (MB)↓ & FPS↑  \\
    \midrule
    D3DGS       & 23.75	& 0.847	& \textbf{0.179}	& 59.38	& 58   \\
    4DHexPlane    & 21.08	& 0.729	& 0.281	& 66.37	 & 83   \\
    4DGS        & 23.00	& 0.814	& 0.259	& 235.95 & 208  \\
    \midrule
    Ours        & \textbf{24.18}	& \textbf{0.855}	& 0.184	& \textbf{0.704}	 & \textbf{274}  \\
  \bottomrule
  \end{tabular}
  \vspace{-4mm}
\end{table}

\subsection{Results}

\paragraph{Rate-distortion performance.}
By adjusting $\lambda_{\rm rate}$, we generate a series of compressed dynamic 3DGS representations under different bitrates. Given a test view and time step, we deform the Gaussian points in the canonical space and render an image, which is then compared to the corresponding ground-truth image. The image quality is evaluated using PSNR, SSIM, and LPIPS metrics. Meanwhile, the size of the compressed dynamic 3DGS representation is recorded as the bitrate. Using bitrate as the x-axis and the three quality metrics as the y-axis, we plot the rate-distortion (RD) curves as shown in Fig.~\ref{fig:scaffold-4dgs_rd}. In addition, we compare the performance of our method against other baselines D3DGS~\cite{yang2024deformable}, 4DHexPlane~\cite{wu20244d} and 4DGS~\cite{yang2024deformable} on the same graph.

As shown in Fig.~\ref{fig:scaffold-4dgs_rd}, on the D-NeRF dataset, our method achieves over 40$\times$ compression with minimal quality degradation compared to D3DGS, mainly benefiting from the efficient spatial prediction and context-aware entropy coding. 4DHexPlane and 4DGS exhibit relatively poor reconstruction quality on synthetic data; even at the lowest bitrate, our method consistently achieves higher objective quality than both. By comparing PSNR between training and testing views, we observe that these two baselines suffer from severe overfitting, showing poor generalization to unseen views.

On the NeRF-DS dataset, our method surpasses the other three baselines in objective quality under high-bitrate settings. At equal levels of quality, our approach achieves over 90$\times$ compression compared to D3DGS.

\begin{figure}[t]
    \centering
    \includegraphics[width=0.95\linewidth]{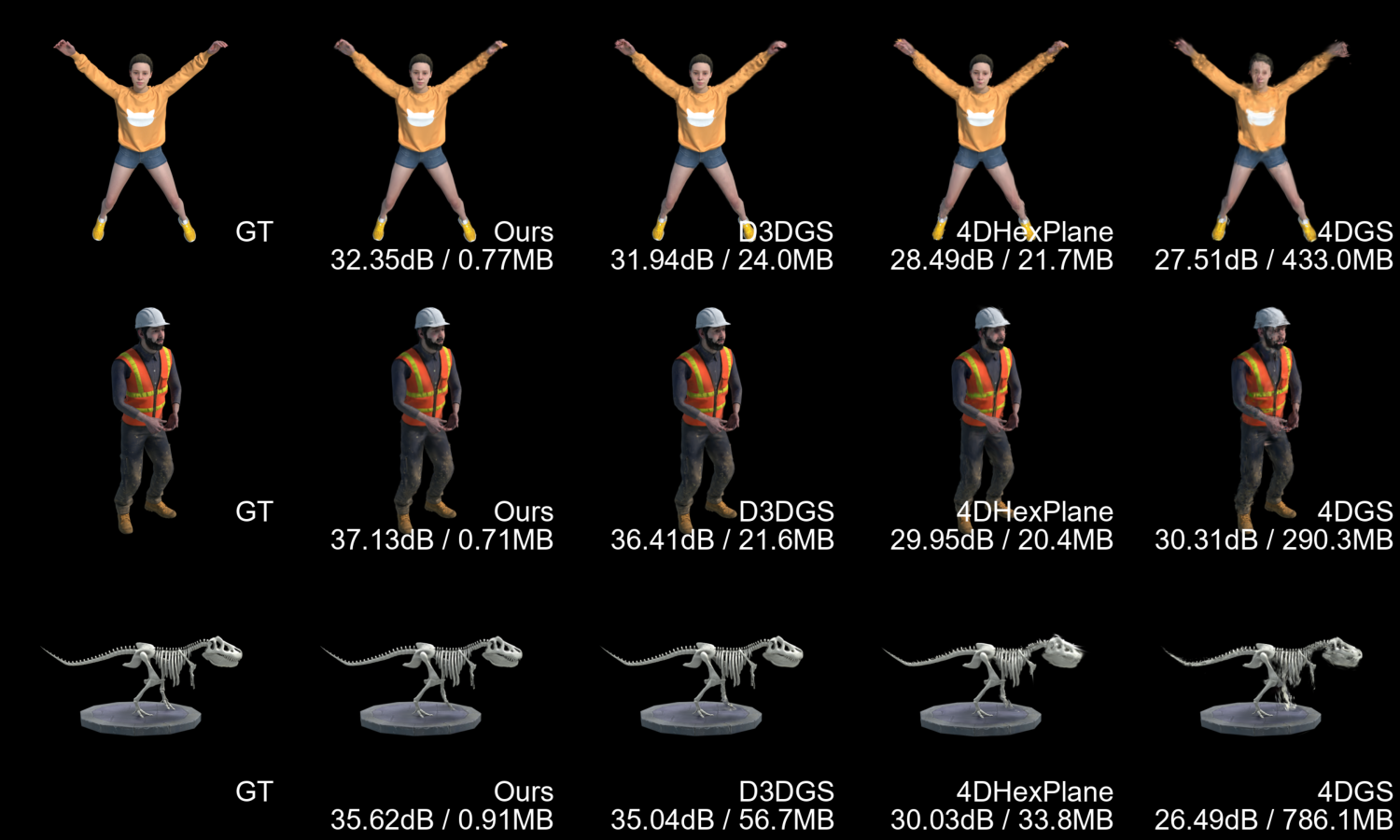}
    \includegraphics[width=0.95\linewidth]{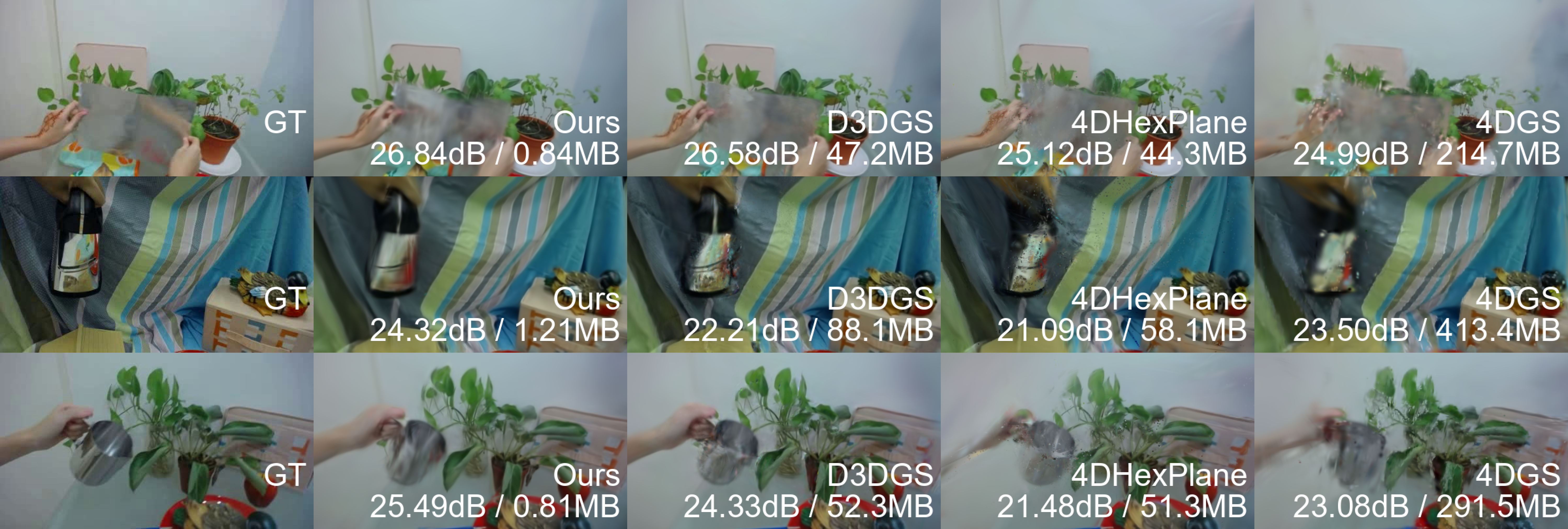}
    \caption{Quantitative comparisons on D-NeRF and NeRF-DS datasets. Our method achieves high-fidelity rendering with significantly lower storage (\textasciitilde1MB) compared to D3DGS, 4DHexPlane, and 4DGS. It faithfully reconstructs dynamic scenes with minimal artifacts, while baseline methods show visible degradation such as blur or loss of dynamic details.}
    \label{fig:perceptual}
    \vspace{-4mm}
\end{figure}

\paragraph{Quantitative comparisons.}
To qualitatively assess the rendering quality of our method compared to other dynamic 3DGS representation approaches, we visualize rendering results from several scenes in both the D-NeRF and NeRF-DS datasets. The comparisons are illustrated in Fig.~\ref{fig:perceptual}. As shown, our method achieves superior rendering quality with minimal storage cost, faithfully reconstructing dynamic 3D scenes with no noticeable artifacts in most cases. In contrast, the baseline methods suffer from various degrees of degradation, including blur, streaks, and visual artifacts, and often fail to reconstruct the dynamic content accurately.

\subsection{Ablation Studies}

\paragraph{Bitrate savings of each module.}
To investigate the contribution of each module to bitrate reduction, we conduct an ablation study on the T-Rex scene. Starting from the baseline model D3DGS (which uses only a temporal prediction module), we incrementally add components from our method. The results are shown in Table~\ref{tab:scaffold-4d bitrate savings}.

The first step is adding the spatial prediction module, which reduces the bitrate from 56 MB to 7 MB (a reduction of approximately 90\%) with an additional 0.3 dB PSNR increase. This improvement can be attributed to the ability of anchor-based prediction to effectively capture spatial redundancy between neighboring Gaussians. Moreover, the number of Gaussians is reduced, enabling the deformation MLP to learn more efficiently and represent the scene with both higher quality and lower storage cost.

The second step involves applying a compact MLP design, which reduces the number of layers in the deformation MLP from 8 to 3 and the feature dimension from 256 to 192. The network parameters are also quantized to \texttt{float16} precision, significantly reducing the model size. This optimization yields around 2 MB of bitrate savings.

Finally, the quantization and context entropy coding modules further compress the representation to under 1MB with negligible loss in quality. This is achieved through quantization-aware training and a learned entropy model that captures the underlying distributions of anchor attributes.

\begin{table}[t]
  \caption{Bitrate savings achieved through various techniques. Beginning with D3DGS~\cite{yang2024deformable}, we incrementally introduce our methodologies to quantify the reduction in bitrate compared to the preceding step.}
  \label{tab:scaffold-4d bitrate savings}
  \centering
  \begin{tabular}{@{}l|ccc@{}}
    \toprule
    Module       & Rate (MB) & Savings (\%) & PSNR  \\
    \midrule
    D3DGS                    & $56.7$   &      -      & $38.04$  \\
    + Spatial Prediction                   &  $7.27$ & $-87.1\%$ & $38.36$\\
    + Compact Deformation MLP                    & $5.39$   &  $-25.8\%$ & $38.34$ \\
    + Quantization \& Entropy Coding               & $0.85$   &  $-84.1\%$ & $38.24$ \\
  \bottomrule
  \end{tabular}
\end{table}

\paragraph{Representation composition.}
We also analyze the component-wise composition of the compressed representation by examining two bitrate settings from the T-Rex scene. The compressed representation consists of anchor attributes (including position, features, scale, and Gaussian offsets), the hash grid, the deformation MLP, and other MLPs such as those for anchor prediction and entropy estimation. Across different bitrates, we observe that the anchor positions and deformation MLP parameters remain constant, as they are quantized with fixed 16-bit precision. In contrast, the sizes of the other components vary with the bitrate, primarily due to changes in the quantization step size introduced by the adaptive quantization strategy.

\begin{table}[t]
  \centering
  \caption{Composition of our high-rate representation and low-rate one trained on scene T-Rex.}
  \label{tab:scaffold4d_repr_comp}
  \begin{tabular}{@{}l@{\hspace{4mm}}|@{\hspace{4mm}}c@{\hspace{4mm}}|@{\hspace{4mm}}c@{\hspace{4mm}}|@{\hspace{4mm}}c@{\hspace{4mm}}|@{\hspace{4mm}}c@{}}
    \toprule
     &
    \multicolumn{2}{c@{\hspace{4mm}}|@{\hspace{4mm}}}{Low rate} &
    \multicolumn{2}{c}{High rate} \\
    \cmidrule{1-5}          Component   & Rate (MB)  & Proportion (\%)   & Rate (MB) & Proportion (\%)   \\
    \midrule
    Anchor position      & $0.121$   & $23.9\%$    & $0.118$    & $13.2\%$       \\
    Anchor feature      & $0.021$   & $4.2\%$     & $0.191$    & $21.4\%$       \\
    Anchor scale & $0.058$   & $11.4\%$     & $0.099$    & $11.1\%$       \\
    Gaussian offsets   & $0.024$   & $4.7\%$    & $0.188$    & $21.1\%$       \\
    Hash grid       & $0.003$   & $0.7\%$    & $0.018$    & $2.0\%$       \\
    Deformable MLP       & $0.241$   & $47.6\%$     & $0.241$   & $26.9\%$       \\
    Other MLPs       & $0.038$   & $7.5\%$     & $0.038$   & $4.2\%$       \\
    \midrule
    Total         & $0.506$  & -             & $0.857$   & -            \\
  \bottomrule
  \end{tabular}
\end{table}

\subsection{Rendering efficiency}
To evaluate the rendering efficiency of our approach, we compare the rendering speed (FPS) of our method with other baselines on two dynamic 3D scene datasets. The results are shown in the last columns of Table~\ref{tab:dnerf_results} and Table~\ref{tab:nerfds_results}. All FPS measurements are conducted on an NVIDIA RTX 4090 GPU. As observed, our method achieves the highest rendering FPS on both datasets.
\section{Conclusion}

In this paper, we investigate an efficient representation tailored for dynamic 3D scenes. Inspired by video encoding techniques, we propose a compact dynamic 3DGS representation framework that integrates spatial-temporal prediction, adaptive quantization and context-based entropy coding. Experimental results demonstrate that our method achieves optimal reconstruction quality and the fastest rendering speed with minimal storage overhead. Compared to the baseline model, our approach attains up to 40× and 90× compression rates on synthetic and real scenes respectively, while maintaining comparable visual quality. The main limitation of our method lies in the deformation MLP, whose storage overhead cannot be adaptively scaled to different bitrates, thereby limiting compression efficiency under low-rate settings. In future work, we plan to explore more compact and compressible temporal representations to address this issue and improve rate adaptability.
\newpage
\appendix

\setcounter{section}{0}
\setcounter{equation}{0}
\setcounter{figure}{0}
\setcounter{table}{0}

\section{Supplementary Material}

In the supplementary material, we provide a detailed description of our hyperparameters in Sec.~\ref{sec:hyperparameters} and more detailed quantitative results in Sec.~\ref{sec:quantitative results}.

\subsection{Hyperparameters}
\label{sec:hyperparameters}

\paragraph{Spatial prediction module.} For the spatial prediction module, anchor points are initialized either using the point cloud estimated by COLMAP (for real-world scenes) or through random sampling in space (for synthetic scenes). The initialized point cloud is voxelized using a voxel size of $\epsilon=0.01$ for real-world scenes or $\epsilon=0.001$ for synthetic scenes to obtain the initial anchor points. Each anchor point can produce up to $k=10$ Gaussian primitives during rendering. As the training proceeds and Gaussians are pruned based on opacity, the value of $k$ gradually decreases, averaging 3 to 4 Gaussians per anchor by the end of training. The update init factor for anchor growing is set to 16. The dimension of anchor attributes is set to $d=16$ for synthetic scenes and $d=32$ for real-world scenes. All anchor MLPs $\psi_{\alpha}$, $\psi_c$, $\psi_r$, and $\psi_s$ consist of 2 hidden layers with dimensions of $2d$.

\paragraph{Temporal prediction module.} The deformation MLP $\psi_d$ consists of two hidden layers, each with 192 dimensions. The output of the final hidden layer is fed into three independent output layers to generate deformation vectors for the position, scale, and rotation attributes of $k$ Gaussian primitives.

\paragraph{Quantization and entropy coding module.} For the quantization and entropy coding module, the base quantization steps for anchor attributes, scale, scaling, and offset are set as $Q_f = 1$, $Q_s = 0.001$, $Q_l = 0.001$, and $Q_o = 0.2$, respectively. Anchor positions are quantized using 16-bit uniform quantization. Both MLPs $\psi_p$ and $\psi_q$ consist of two hidden layers, each with a hidden dimension of $2d$.

\paragraph{Learning rate settings.} For anchor attributes, the learning rate is set to 0.01 for offsets, 0.0075 for anchor feature, 0.02 for opacity, and 0.007 for scale and offset scaling. For anchor MLPs, the learning rates are 0.004 for covariance, 0.008 for color, and 0.002 for opacity. The deformation MLP is trained with a learning rate of 0.00016. The learning rates of hash grid and MLPs related to quantization and entropy coding are set to 0.0016 and 0.005, respectively. During training, learning rates decay exponentially with increasing iterations.

\subsection{Quantitative Results}
\label{sec:quantitative results}
In this section, we provide detailed experimental results, including per-scene performance comparisons between our method and three baseline approaches (Tab.\ref{tab:quantitative comparison per scene on D-NeRF dataset} and Tab.\ref{tab:quantitative comparison per scene on NeRF-DS dataset}). Additionally, we report the performance of our method under varying compression rates on two datasets (Tab.~\ref{tab:Quantitative comparison of our method with different rates on D-NeRF and NeRF-DS datasets}).

\begin{table}[H]
  \caption{Per-scene quantitative comparison on D-NeRF dataset. We acquire the baseline results by running their official codes.}
  \label{tab:quantitative comparison per scene on D-NeRF dataset}
  \centering
  \begin{tabular}{l|l|cccc}
    \toprule
    &Scene&bouncingballs&hellwarrior&hook&jumpingjacks \\
    \midrule
    \multirow{4}{*}{D3DGS}
    & Rate (MB) &45.27  &10.99  &38.99  &24.04    \\
    & PSNR      &40.56  &41.35  &36.98  &37.57    \\
    & SSIM      &0.995  &0.987  &0.986  &0.990    \\
    & LPIPS     &0.010  &0.024  &0.016  &0.013    \\
    \midrule
    \multirow{4}{*}{4DHexPlane} 
    & Rate (MB) &19.70  &21.22  &20.97  &20.78    \\
    & PSNR      &40.74  &28.68  &32.93  &35.34    \\
    & SSIM      &0.994  &0.973  &0.977  &0.986    \\
    & LPIPS     &0.015  &0.037  &0.027  &0.020    \\
    \midrule
    \multirow{4}{*}{4DGS}
    & Rate (MB) &243.19 &400.54 &322.34 &433.92   \\
    & PSNR      &32.73  &34.24  &31.54  &32.24   \\
    & SSIM      &0.983  &0.955  &0.959  &0.971   \\
    & LPIPS     &0.030  &0.080  &0.047  &0.037   \\
    \midrule
    \multirow{4}{*}{Ours} 
    & Rate (MB) &1.07   &0.63   &1.33   &0.73      \\
    & PSNR      &42.41  &40.79  &36.49  &37.70     \\
    & SSIM      &0.996  &0.987  &0.985  &0.990     \\
    & LPIPS     &0.008  &0.027  &0.017  &0.015     \\
    \bottomrule
    \toprule
    &Scene&lego&mutant&standup&trex \\
    \midrule
    \multirow{4}{*}{D3DGS}
    & Rate (MB) &73.53 &44.44  &21.64  &56.72   \\
    & PSNR      &24.92 &42.63  &44.24  &38.04   \\
    & SSIM      &0.943 &0.995  &0.995  &0.993   \\
    & LPIPS     &0.044 &0.005  &0.007  &0.010   \\
    \midrule
    \multirow{4}{*}{4DHexPlane} 
    & Rate (MB) &33.02 &22.45  &19.42  &31.83   \\
    & PSNR      &25.05 &37.71  &38.06  &33.69   \\
    & SSIM      &0.938 &0.988  &0.990  &0.984   \\
    & LPIPS     &0.056 &0.016  &0.014  &0.022   \\
    \midrule
    \multirow{4}{*}{4DGS}
    & Rate (MB) &423.43&301.36 &347.14 &751.50  \\
    & PSNR      &24.23 &36.62  &38.29  &29.93  \\
    & SSIM      &0.908 &0.982  &0.985  &0.976  \\
    & LPIPS     &0.096 &0.019  &0.017  &0.029  \\
    \midrule
    \multirow{4}{*}{Ours} 
    & Rate (MB) &2.04 &0.93   &0.70   &0.88      \\
    & PSNR      &24.57&41.12  &43.53  &38.25     \\
    & SSIM      &0.939&0.993  &0.994  &0.993     \\
    & LPIPS     &0.049&0.009  &0.008  &0.011     \\
  \bottomrule
  \end{tabular}
\end{table}

\begin{table}[H]
  \caption{Per-scene quantitative comparison on NeRF-DS dataset. We acquire the baseline results by running their official codes.}
  \label{tab:quantitative comparison per scene on NeRF-DS dataset}
  \centering
  \begin{tabular}{l|l|cccccccc}
    \toprule
    & Scene &as&basin&bell&cup&plate&press&sieve \\
    \midrule
    \multirow{4}{*}{D3DGS}
    & Rate (MB)&47.24&68.67&88.15&52.32&57.22&49.09&52.99 \\
    & PSNR&25.91&19.64&25.06&24.74&20.14&25.37&25.45 \\
    & SSIM&0.881&0.788&0.841&0.889&0.806&0.859&0.872 \\
    & LPIPS&0.183&0.189&0.164&0.155&0.221&0.193&0.150 \\
    \midrule
    \multirow{4}{*}{4DHexPlane}
    & Rate (MB)&52.36&60.16&65.58&52.04&103.00&60.07&50.51 \\
    & PSNR&23.25&18.65&21.50&22.66&16.38&20.88&24.28 \\
    & SSIM&0.775&0.704&0.729&0.833&0.613&0.643&0.812 \\
    & LPIPS&0.258&0.264&0.283&0.187&0.418&0.368&0.195 \\
    \midrule
    \multirow{4}{*}{4DGS}
    & Rate (MB)&204.09&128.75&396.73&280.38&124.93&275.61&164.99 \\
    & PSNR&24.97&19.20&23.96&23.74&19.74&25.54&23.84 \\
    & SSIM&0.845&0.750&0.811&0.854&0.786&0.838&0.815 \\
    & LPIPS&0.248&0.299&0.236&0.219&0.295&0.263&0.260 \\
    \midrule
    \multirow{4}{*}{Ours} 
    & Rate (MB)&0.67&0.66&0.90&0.67&0.66&0.75&0.62 \\
    & PSNR&26.72&19.95&25.38&25.07&21.02&25.76&25.37 \\
    & SSIM&0.885&0.802&0.845&0.896&0.826&0.864&0.871 \\
    & LPIPS&0.181&0.204&0.174&0.157&0.211&0.202&0.160 \\
  \bottomrule
  \end{tabular}
\end{table}

\begin{table}[H]
  \caption{Performance of our method under varying bitrates.}
  \label{tab:Quantitative comparison of our method with different rates on D-NeRF and NeRF-DS datasets}
  \centering
  \begin{tabular}{l|cccc|cccc}
    \toprule
    Dataset & \multicolumn{4}{c|}{D-NeRF} & \multicolumn{4}{c}{NeRF-DS} \\
    \midrule
    Rate (MB)& 1.03&0.78&0.71&0.57&0.70&0.65&0.57&0.54 \\
    PSNR& 38.10&37.73&37.38&35.10&24.18&23.93&23.60&23.34 \\
    SSIM& 0.984&0.984&0.984&0.978&0.856&0.849&0.832&0.822 \\
    LPIPS& 0.018&0.018625&0.0195&0.031&0.184&0.204&0.246&0.267 \\
    \bottomrule
  \end{tabular}
\end{table}

\bibliographystyle{unsrt}
\bibliography{neurips_2025}

\end{document}